\documentclass{ecai}
\usepackage{graphicx}
\usepackage{latexsym}

\usepackage{algorithm}
\usepackage{algorithmic}

\usepackage{amsthm}
\newtheorem{definition}{Definition}

\usepackage{amsfonts,amssymb}
\usepackage{comment}
\usepackage{booktabs}
\usepackage{dsfont}
\usepackage{listings}
\usepackage{amsmath,amssymb}
\usepackage{amsthm}
\usepackage{comment}

\begin{document}

\begin{frontmatter}

\title{Associative Memories in the Feature Space}

\author[A]{\fnms{Tommaso}~\snm{Salvatori}~\orcid{0000-0002-7254-9882}\thanks{Corresponding Author. Email: tommaso.salvatori@verses.ai. This work was presented at the 26th European Conference on Artificial Intelligence, 2023. }}
\author[B]{\fnms{Beren}~\snm{Millidge}}
\author[B]{\fnms{Yuhang}~\snm{Song}}
\author[B]{\fnms{Rafal}~\snm{Bogacz}}
\author[E,D]{\fnms{Thomas}~\snm{Lukasiewicz}~\orcid{0000-0002-7644-1668}}

\address[A]{VERSES AI Research Lab, Los Angeles, CA 90016, USA}
\address[B]{Medical Research Council Brain Network Dynamics Unit, University of Oxford, UK}
\address[E]{Institute of Logic and Computation, Vienna University of Technology, Austria}
\address[D]{Department of Computer Science, University of Oxford, UK}

\begin{abstract}
An autoassociative memory model is a function that, given a set of data points, takes as input an arbitrary vector and outputs the \emph{most similar} data point from the memorized set. However, popular memory models fail to retrieve images even when the corruption is mild and easy to detect for a human evaluator. This is because similarities are evaluated in the raw pixel space, which does not contain any semantic information about the images. This problem can be easily solved by computing \emph{similarities} in an embedding space instead of the pixel space. We show that an effective way of computing such embeddings is via a  network pretrained with a contrastive loss. As the dimension of embedding spaces is often significantly smaller than the pixel space,  we also have a faster computation of similarity scores. We test this method on complex datasets such as CIFAR10 and STL10. An additional drawback of current models is the need of storing the whole dataset in the pixel space, which is often extremely large. We relax this condition and propose a class of memory models that only stores low-dimensional semantic embeddings, and uses them to retrieve similar, but not identical, memories. We demonstrate a proof of concept of this method on a simple task on the MNIST dataset.
\end{abstract}

\end{frontmatter}

\section{Introduction}

Throughout life, our brain stores a huge amount of information in memory, and can flexibly retrieve memories based on related stimuli. This ability is key to being able to perform intelligently on many tasks. In the brain, sensory neurons detect external inputs and transmit this information to the hippocampus via a hierarchical network, which can retrieve in a constructive way via a generative network \cite{Barron20}. Stored memories that involve a conscious effort to be retrieved  are called \emph{explicit}, and are divided into  \emph{episodic} and \emph{semantic} memories. Episodic memories consists of experienced events, while semantic memories represent knowledge and concepts. Both these memories are retrieved in a constructive way via a generative network \cite{Barron20}.

In computer science, computational models of associative memories are basically pattern storage and retrieval systems. A standard task is to store a dataset, and retrieve the correct data point when shown a corrupted version of it \cite{Hopfield82,Hopfield84}. Popular associative memory models are Hopfield networks \cite{Hopfield82,Hopfield84}, with their modern continuous state formulation \cite{ramsauer21,krotov16}, and sparse distributed memories \cite{kanerva1988sparse}. While these models have a large theoretical capacity, which can be exponential in the case of continuous-state Hopfield networks \cite{ramsauer21,Krotov21}, this is not reflected in practice, as they fail to correctly retrieve memories such as high-quality images when presented with even medium-size datasets \cite{millidge2022universal,salvatori2021associative}. In fact, the similarity between two points is typically computed on the raw pixel space using a simple function (such as a dot product) that is insensitive to the `semantic' features of images that we wish to discriminate between.  The performance would drop even more when using stronger corruptions, such as rotations, croppings, and translations, as relations between individual pixels would be lost. These problems can be solved by learning a similarity function that is sensitive to the semantics of the stored memories. In essence, we need to embed every data point into a different space, where simple similarity scores can discriminate well between semantic features. This approach resembles kernel methods, where the similarity operation is performed after the application of a feature map $\phi$, which sends both the input and the data points to a space where the dot product is more meaningful.

\begin{figure*}[t]
\includegraphics[width=1\textwidth]{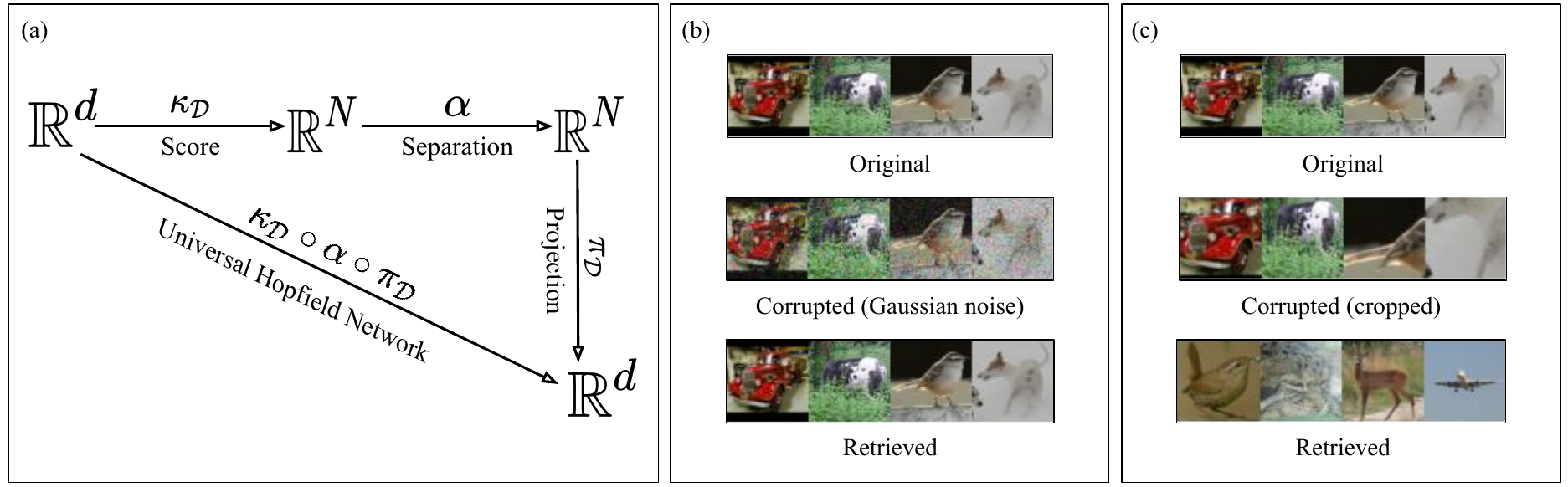} 
\caption[short]{(a): Decomposition of a universal Hopfield network in score, similarity, and projection. (b)~Examples of retrieved data points when given corrupted versions using Gaussian noise.  (c)~Examples of retrieved data points when given cropped versions.  }
\label{fig:shn_intro}\vspace*{-0.5ex}\end{figure*}

This leads to the problem of finding a map $\phi$ that embeds different data points in a space where they can all be well discriminated. In this work, we demonstrate that the simple approach of using pre-trained neural networks as feature maps strongly improves the performance of standard Hopfield networks. We first review a recent mathematical formalism that describes one-shot associative memory models present in the literature, called \emph{universal Hopfield networks}, and extend this framework to incorporate these features maps. The main contributions of this paper are briefly as follows:
\begin{itemize}

    \item We define a class of associative memory models, called \emph{semantic Hopfield networks}, that augment associative memory models with a feature map. In this case, as a feature map, we use ResNet18 and ResNet50, pretrained in a contrastive way, as done in SimCRL~\cite{chen2020simple}. What results is a model that stores the original data points as in standard memory models, but computes similarities in an embedding space. This model is able to perform an exact retrieval on complex data points, such as CIFAR10, STL10, and ImageNet images, when presented with queries formed by corrupted and incomplete versions.

    \item We then address another drawback of current associative memory models, namely, the need to store all data points, which is memory-inefficient. To this end, we propose a model that stores low-dimensional embeddings of the original data points. The retrieved data points are not exact copies of the stored ones, as they are generated via a generative network $\psi: \mathbb{R}^k \longrightarrow \mathbb{R}^d$. This also adds a degree of biological plausibility, as the data points in this model are stored in a declarative way, and retrieved in a constructive way. We provide a proof of concept of this model on MNIST, using a simple autoencoder.

\end{itemize}

The rest of this paper is structured as follows. In Section 2, we introduce Universal Hopfield networks, providing formal definitions that describe their structure. Sections 3 and 4 introduce the original contributions of this work, the \emph{semantic memory model} and its fully-semantic variation. In Sections 5 and 6, we end the paper with a summary of the related literature and a conclusive discussion. 

\section{Preliminaries}

In this section, we review  \emph{universal Hopfield networks} \cite{millidge2022universal}. According to this framework, associative memory models can always be represented as decompositions of three parametrized functions: \emph{score, separation}, and \emph{projection}, whose parameters depend on the stored memories. Let $\mathcal{D} = \{\bar x_i\}_{i \leq N}$ be a dataset, with $\bar x_i \in \mathbb{R}^d$ for every $i$. Informally, given any  $\bar x \in \mathbb{R}^d$, the goal of an associative memory model is to return the data point of $\mathcal D$ that is \emph{most similar} to $\bar x$ according to a function $\kappa:\mathbb R^d \times \mathbb R^d \longrightarrow \mathbb{R}$. Hence, we have the following:
\begin{definition}
{\rm Given a dataset $\mathcal{D} = \{\bar x_i\}_{i \leq N}$, a \emph{universal Hopfield network} is a function ${\mu_{\mathcal{D}}: \mathbb{R}^d \longrightarrow \mathbb{R}^d}$ such that $\mu_{\mathcal{D}}$ admits a decomposition $\mu_{\mathcal{D}} = \pi_{\mathcal{D}} \circ \alpha \circ \kappa_{\mathcal{D}} $ into:
\begin{enumerate}
    \item \textit{score:} a function $\kappa_{\mathcal{D}}:\mathbb R^d \longrightarrow \mathbb R^N$ such that $\kappa_{\mathcal{D}}(\bar x)_i = \kappa(\bar x, \bar x_i)$,
    \item \textit{separation:} a function $\alpha: \mathbb R^N \longrightarrow \mathbb R^N$ not dependent on the dataset,
    \item \textit{projection:} a function $\pi_{\mathcal{D}}: \mathbb R^N \longrightarrow \mathbb R^d$ dependent on the dataset.
\end{enumerate}
}\end{definition}
Ideally, we would like the function $\mu_{\mathcal{D}}$ to store the dataset $\mathcal{D}$ as an attractor of its dynamics. Informally, an attractor is a set of points that a system tends to evolve towards. 
In designing associative memories, we typically wish to store data points as attracting points and design a retrieval function $f$ that converges to the data points in as few iterations as possible. In a continuous space, however, the attractors may be close to the data points, but not exactly where the data points are. This depends on the choice of the separation function. However, this problem is easily solved by taking the maximum value after computing the separation function. In practice, models able to retrieve data points in one shot are preferable. This is always the case when using \emph{max} as separation function, or a continuous approximation given by a softmax with a large inverse-temperature $\beta$. We now describe the main ideas behind the decomposition of universal Hopfield networks, and show how popular models in the literature can be derived from~it.

 \paragraph{Score.} Given an input vector $\bar x$, the score returns a vector that has the number of entries equal to the number of data points $N$. The $i$-th entry of the vector $\kappa_{\mathcal{D}}(\bar x)$ represents how similar $\bar x$ is relative to the data point $\bar x_i$. Hopfield networks compute the similarity using a dot product, while sparse distributed memories use the negative Hamming distance.

\paragraph{Separation.} If the cardinality of the dataset is large, and multiple data points are close to the input $\bar x$ in terms of similarity, the retrieval process may require a large number of iterations of $\mu$. However, we wish to retrieve a specific data point as quickly as possible. The goal of the separation function $\alpha$ is then to emphasize the top score and de-emphasize the rest, to make convergence faster. Popular choices of separation functions are \emph{softmax}, \emph{threshold}, and \emph{polynomial}, used respectively by modern Hopfield networks, sparse distributed memories, and dense Hopfield networks \cite{krotov16,Demircigil17}. 


 \paragraph{Projection.} The projection is a function that, given the vector with scores, already modified by the separation function, returns a vector in the original input space.  For exact retrieval, the projection function is set to the matrix of data points. Particularly,  consider the matrix $P \in \mathbb{R}^{d \times N}$, which has its i-th column equal to the data point $\bar x_i$. Then, we have $\pi( \bar x) = P \bar x$. If $\bar x$ is a $1$-hot vector, a perfect copy of a data point is returned.

This new categorization of one-shot memory models has enabled a systematic testing and generalization over multiple combinations of similarity functions, showing that, for image datasets, some similarity and separation functions work much better than others. For example, metrics such as negative $L1$ and $L2$ distances outperform dot products. However, distances are less biologically plausible than dot products, as they require special computations to be computed, while a dot product can be represented as a perceptron layer. As shown in Fig.~\ref{fig:shn_intro}, scoring images on the pixel space is highly impractical, as it suffices to simply  rotate or crop an image to trick the memory model. For separation functions, we use softmax with a large inverse-temperature $\beta$, as it is able to approximate the max function and hence perform one-shot retrievals.

\section{Semantic Memory Model}

In this section, we propose a new class of associative memory models. Intuitively, this class is similar to UHNs, but is augmented with an embedding function $\phi$ that maps memories into a feature space. Here, two embeddings are scored as in UHNs as if they were the original stored data points. The resulting vector with the similarity scores is first separated, and then projected back to the pixel space. We will show that this approach enables powerful associative memory models. 

\begin{definition}
Given a dataset $\mathcal{D} = \{\bar x_i\}_{i \leq N}$ and a feature map $\phi: \mathbb{R}^d \longrightarrow \mathbb{R}^e$, a \emph{semantic memory model} is a function ${\mu_{\mathcal{D}}: \mathbb{R}^d \longrightarrow \mathbb{R}^d}$ such that:
\begin{enumerate}
    \item $\mu_{\mathcal{D}}$ admits the decomposition $\mu_{\mathcal{D}} =  \pi_{\mathcal{D}} \circ \alpha \circ \kappa_{\phi(\mathcal{D})} \circ \phi  $, 

    \item the map $ \pi_{\mathcal{D}} \circ \alpha \circ \kappa_{\phi(\mathcal{D})}$ is a universal Hopfield network, where similarity scores are computed in the embedding space $\mathbb{R}^e$.
\end{enumerate} 
\end{definition}

\begin{figure*}[t]
\includegraphics[width=1\textwidth]{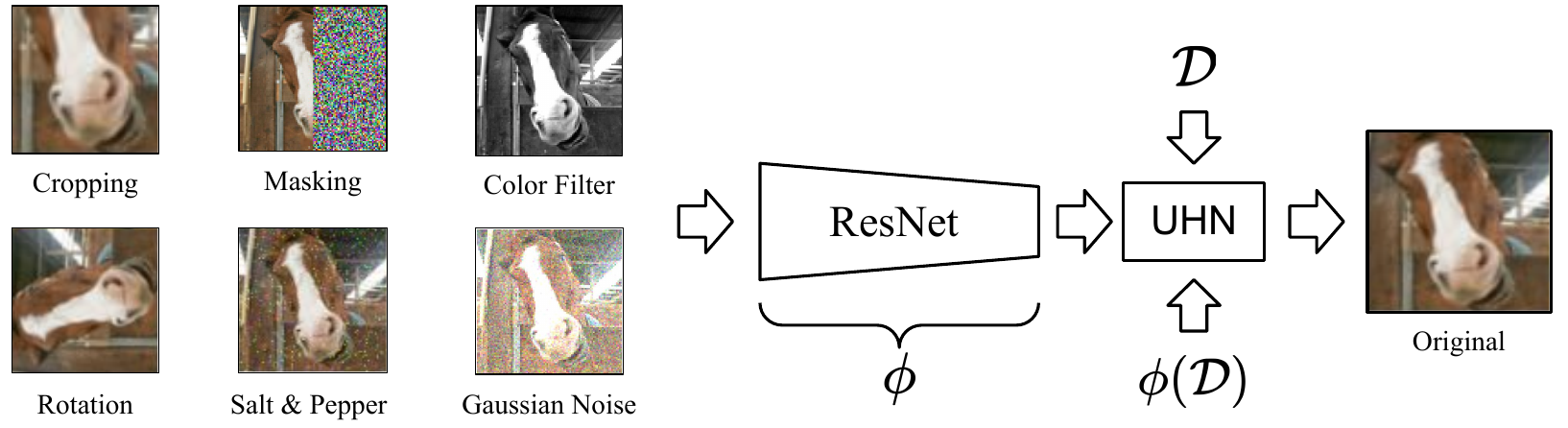}
\caption[short]{ Example of a semantic memory model, where the function $\phi$ is a ResNet pre-trained using a contrastive loss. On the left, examples of the six kinds of corruptions used in this section; on the right, the original image to be retrieved by the model.}
\label{fig:qshn}
\end{figure*}

\paragraph{SimCLR.} The first problem to address is to find a suitable embedding $\phi$ to perform associative memory experiments. Ideally, this function should map corrupted versions of the same data point close to each other, and different data points away from each other. A straightforward way of doing this is to train a neural network using a contrastive loss. This has already been done in the literature, as it is an effective way of pre-training a neural network when a large amount of unlabelled data is available \cite{simclr,chen2020big}. Typically, the pre-training procedure works as follows: given a dataset $\mathcal D = \{\bar x_i\}_{i \leq N}$, the network is provided simultaneously with a batch of $B$ pairs of data points $\tilde x_i,\tilde x_j$ that are corrupted versions of the data points $\bar x_i, \bar x_j$, and trained to minimize the contrastive loss: 
\begin{align*}
    \mathcal L_{i,j} = -log ( \frac{exp(sim(\tilde z_i,\tilde z_j))}{{\sum\nolimits^{2B}_{k=0} \mathds{1}_{i \neq k}exp(sim(\tilde z_i,\tilde z_k))}}),
\end{align*}
where $\tilde z_i = \phi(\tilde x_i)$ is the output of the network, $\mathds{1}_{i,k}$ is a binary function equal to one, if $i \neq k$, and zero, otherwise, and \emph{sim} is a similarity function. When training has converged, the original work then adds a feedforward layer (or more) attached to the output layer, where the contrastive loss is defined, to fine-tune using the few labelled data available. This simple framework for contrastive learning of visual representations is known as \emph{SimCRL}. As we do not need to perform supervised learning, here we simply use the pretrained network to compute similarity scores of pairs of data points embedded into the latent space of the model.

\paragraph{Set-up.} In the following experiments, we test our semantic memory model on two datasets, CIFAR10 \cite{Krizhevsky2012} and STL10 \cite{stl10}. The first one consists of $60000$ $32 \times 32$ colored images, divided in a $50000-10000$ train-test split, while the second consists of $105000$ $96 \times 96$ colored images, divided in a $100000-5000$ train-test split. As functions $\phi$, we use a ResNet18 for CIFAR10  and a ResNet50 for STL10 \cite{he2016deep}, trained as described in the original SimCLR work \cite{simclr}. Details about the parameters used can be found in the supplementary material. Then, we use the test sets, never seen by the models, to evaluate the retrieval performance from corrupted memories. Particularly, we use the following six kinds of corruptions, visually explained on the left side of Fig.~\ref{fig:qshn}:
\begin{enumerate}
    \item \textbf{Cropping (Crop)}: the corrupted image is a zoomed version of the original one, 
    
    \item \textbf{Masking (Mask)}: half of the image is masked with uniform random noise, 
    
    \item \textbf{Color filters (Color)}: different color filters are randomly applied to the original images, 
    
    \item \textbf{Rotation}: the images are randomly rotated by an angle of $0,\pi/2,-\pi/2,\pi$,

    \item \textbf{Salt and pepper (S\&P)}: a random subset of the pixels of the original images is set to 1 or 0, 
    
    \item \textbf{Gaussian noise (Gauss)}: Gaussian noise of variance $\eta = 0.1$ and different means is added to the original images.
    
\end{enumerate}
As similarity functions, we  tested the dot product, the cosine similarity, the negative Euclidean distance (L2 norm), and the negative Manhattan distance (L1 norm). As separation function, we used a softmax with large inverse temperature. To make the comparison with UHNs clear, we also report the accuracies using the same corruptions and activation functions.

\paragraph{Implementation Details.} As a loss function, we always used a contrastive loss with cosine similarity, as done in the original work on SimCLR. As parameters, we followed a popular PyTorch implementation.\footnote{https://github.com/sthalles/SimCLR} It differs from the official one, which is only available in TensorFlow, but is equivalent in terms of the pre-training  regime. For the experiments on CIFAR10, we used a ResNet18 with embedding dimension $512$ trained for $100$ epochs; for STL10, we  used a ResNet50 with embedding dimension $2048$ trained for $50$ epochs. The hyperparameters used for both models are the same: batch size of $256$, learning rate of $0.0003$, and weight decay of $1e-4$. As it is complex to exactly describe the details of the corruptions used to perform our associative memory tasks, we refer to the {PyTorch} code in the supplementary material. For the first three corruptions, rotations, filters, and croppings, we  have used the relative torchvision transformations. For Gaussian noise, masks, and salt and pepper noise, we  report the corruption on the original data point. The following code allows to generate the same corruptions of Fig.~2.

\begin{table*}[t]
    \caption{Percentage of wrongly retrieved memories on CIFAR10.}
    \centering{\vspace*{-1ex}
    	\resizebox{\textwidth}{!}{\begin{tabular}{@{}lcccccccc@{}}
    		\toprule
    		& Crop  & Mask & Color & Rotation & S\&P & Gauss $\mu =0.3$ & Gauss $\mu = 0.5$    \\
    		\toprule
    		 
    		 UHN (Cosine Sim.)  & $74.53\%$  & $72.11\%$ & $19.36\%$ & $42.73\%$ & $99.12\%$ & $29.12\%$ &$99.19\%$ \\
    		 
    		 UHN (L2 Norm)  & $65.73\%$  & $37.42\%$ & $51.31\%$ & $39.91\%$  & $4.31\%$  & $85.13\%$ & $99.92\%$ \\
    		 
    		 UHN (L1 Norm)  & $57.32\%$  & $\textbf{2.41\%}$ & $57.11\%$ & $33.13\%$  & $4.17\%$  & $91.11\%$ & $99.81\%$ \\ \\

    		 Ours (Cosine Sim.)  & $\textbf{25.34\%}$  & $63.69\%$ & $0.09\%$ & $\textbf{0.05\%}$ & $20.50\%$ & $15.38\%$ &$52.11\%$ \\
    		 
    		 Ours (L2 Norm)  & $25.89\%$  & $88.03\%$ & $\textbf{0.04\%}$ & $0.06\%$  & $14.31\%$  & $18.78\%$ & $29.83\%$ \\
    		 
    		 Ours (L1 Norm)  & $30.55\%$  & $20.33\%$ & $0.05\%$ & $0.33\%$  & $\textbf{3.65\%}$  &   $\textbf{ 7.19\%}$ & $\textbf{27.88\%}$ \\ 
    		\bottomrule
    	\end{tabular}}}
	\label{tab:results}
\end{table*}

\begin{table*}[t]
    \caption{Percentage of wrongly retrieved memories on STL10.}
    \centering{\vspace*{-1ex}
    		\resizebox{\textwidth}{!}{\begin{tabular}{@{}lcccccccc@{}}
    		\toprule
    		& Crop  & Mask & Color & Rotation & S\&P & Gauss $\mu = 0.3$ & Gauss $\mu = 0.5$    \\
    		\toprule
    		 
    		 UHN (Cosine Sim.)  & $81.12\%$  & $63.77\%$ & $16.39\%$ & $41.11\%$ & $99.57\%$ & $21.37\%$ &$99.98\%$ \\
    		 
    		 UHN (L2 Norm)  & $77.133\%$  & $\textbf{0.09\%}$ & $40.81\%$ & $39.61\%$  & $4.13\%$  & $80.36\%$ & $99.51\%$ \\
    		 
    		 UHN (L1 Norm)  & $65.15\%$  & $0.13\%$ & $33.62\%$ & $31.43\%$  & $16.18\%$  & $88.02\%$ & $99.65\%$ \\ \\

    		 Ours (Cosine Sim.)  & $\textbf{31.12\%}$  & $57.27\%$ & $0.22\%$ & $\textbf{0.03\%}$ & $19.83\%$ & $15.38\%$ &$52.11\%$ \\
    		 
    		 Ours (L2 Norm)  & $34.13\%$  & $51.72\%$ & $\textbf{0.20\%}$ & $0.04\%$  & $31.72\%$  & $18.78\%$ & $49.18\%$ \\
    		 
    		 Ours (L1 Norm)  & $33.32\%$  & $59.11\%$ & $0.36\%$ & $0.08\%$  & $\textbf{3.66\%}$  & $\textbf{6.19\%}$ & $\textbf{17.92\%}$ \\ 
    		 
    		\bottomrule
    	\end{tabular}}}
	\label{tab:results_stl}
\end{table*}

\begin{table}
    \caption{Running times of the experiments (in seconds).}
\centering{\vspace*{-1ex}\begin{tabular}{@{}lcccc@{}}
    		\toprule
    		  & Cosine & L2 Norm & L1 Norm \\
    		\toprule
    		 CIFAR10   & $7.601$ & $8.858$ & $8.904$ \\ 
    		STL10   & $41.788$ & $44.973$ & $47.835$   \\
    		CIFAR10 (ours)   & $2.931$ & $3.282$ & $3.310$ \\ 
STL10 (ours)  & $12.662$ & $14.439$ & $15.646$ \\
    		\bottomrule
    	\end{tabular}}
	\label{tab:times}
\end{table}

\paragraph{Results.} Detailed results about the performance of this method, where the percentage of wrongly retrieved images for each task, dataset, and similarity function are given in Tables~\ref{tab:results} and \ref{tab:results_stl}. As expected, our models outperform UHNs on corruptions where the position of the pixels is altered. This corresponds to all the corruptions considered, besides masking and salt and pepper noise. In fact, when masking an image, $50\%$ of the pixels remain unchanged, allowing similarity functions on the pixel space to return high values. In this task, UHNs outperform our models. A similar reasoning can be applied to salt and pepper noise. Here, however, our method performs better by a small margin. 

In all the other considered tasks, the margin is large, and the few correctly retrieved images by UHNs belong to particular cases: UHNs were able to retrieve cropped or rotated images only when they had close to uniform colors/backgrounds. In those cases, in fact, it is much more likely that a crop or a rotation leaves the embedding of an image in the pixel space mostly unchanged. Uniform images are in fact fixed points of those transformations. 

In terms of similarity functions used, semantic models are generally more robust than UHNs, where the final performance of a specific similarity function strongly depends on the corruption used. In most cases, the cosine similarity and distances obtained a completely different performance. While this also happened in some cases for our model, the negative L1 norm always obtained the best (or close to the best) performance. For UHNs, no similarity has shown to be preferable to the others. This is an advantage of semantic models, as we want to build a memory model that is robust under different kinds of corruptions. 

\paragraph{Efficiency.}
In this paragraph, we show the better efficiency of our method against standard memory models. As already stated, dot products are slightly faster than distances to be computed. However, under some kinds of corruptions, the better performance of the L1 norm makes it the best candidate.  In Table~\ref{tab:times}, we have compared the running times of the proposed experiments. The results show that semantic models are much faster than UHNs, despite the fact that they have to perform a forward pass to compute the semantic embeddings. This better efficiency is simply a consequence of the smaller dimension of the embedding space with respect to the pixel space, but it may be crucial in some scenarios. Particularly, the dimension of the semantic spaces is given by the dimension of the output of the embedding function $\phi$ considered, in our case $512$ for ResNet18 and $2048$ for ResNet50. This is a large improvement over the pixel space, as a single CIFAR10 image has the dimension $3072$ and a single STL10 image has the dimension $27648$. In tasks where having an efficient model is a high priority, it is possible to speed up the model by using pre-trained models with a smaller output dimension. This could be important in online applications. 

\begin{figure*}[t]
\centering{\includegraphics[width=1.0\textwidth]{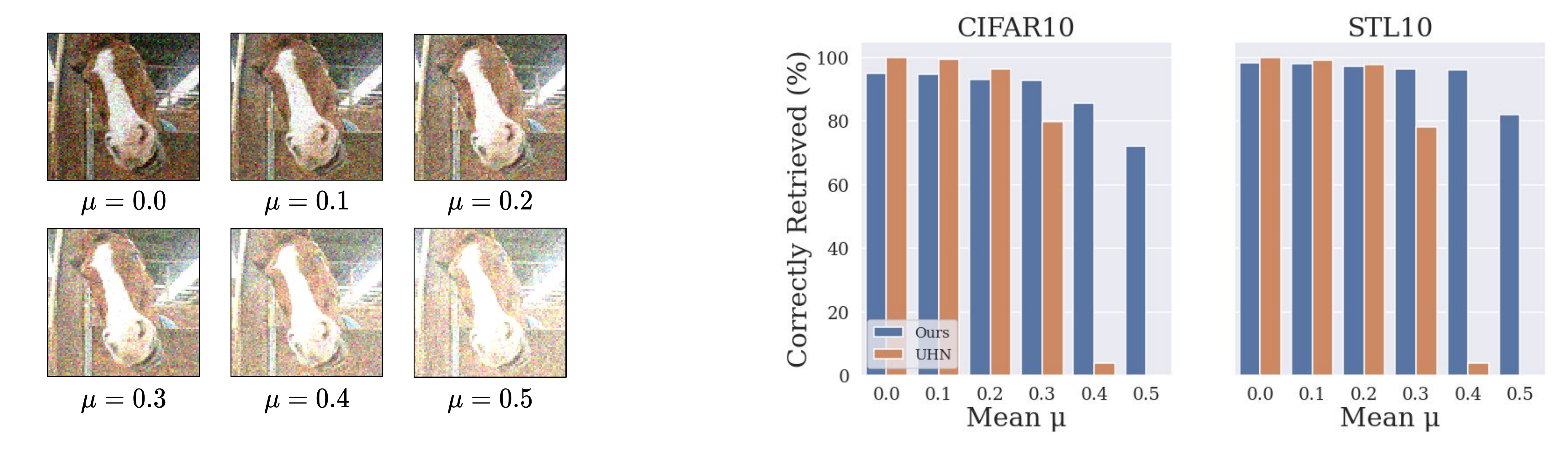}\caption[short]{ Retrieval accuracies of UHNs and semantic models when presented with images corrupted with Gaussian noise of variance $\eta = 0.1$ and different levels of mean $\mu$. On the left, examples of images after this corruption was applied; on the right, retrieval accuracies plotted considering the best result obtained testing different similarity functions. }}\label{fig:gauss_shn}
\end{figure*}

\paragraph{Changing the Mean.} To better study how the two models differ when retrieving images with different levels of noise, we replicate the experiments performed above using as corruptions added Gaussian noise with different means ($\mu=\{0,0.1,0.2,0.3,0.4,0.5\}$), and variance $0.1$. Visual examples of the resulting corrupted images are given on the left side of Fig.~\ref{fig:gauss_shn}. This kind of noise corrupts the image by both adding random noise, and by making it ``whiter''. UHNs are robust with respect to noise with zero mean \cite{millidge2022universal}, but weak when this is increased, as they have a large impact on the position of an image in the pixel space. Making an image ``whiter'', however, does not alter the semantic information that it contains: from a human perspective, we are easily able to determine that the six images represented on the left side of Fig.~\ref{fig:gauss_shn} are different corrupted versions of the same image. Hence, we expect semantic models to perform better than UHNs when dealing with images corrupted by adding Gaussian noises of high mean. This is indeed the case, as the results presented on the right side of Fig.~\ref{fig:gauss_shn} show. Here, the performance of the two models is comparable (with UHNs being slightly better) when using a mean of $0.3$ or smaller. The performance of UHNs, however, significantly dropped when using higher means: they were able to retrieve less than $5\%$ of the images when presented with Gaussian noise with mean $0.4$, and less than $1\%$ when this mean was further increased to $0.5$. Instead, the performance of semantic models were stable, and suffered only a small decrease: they were able to always retrieve more than $70\%$ of the original memories when presented with Gaussian noise of mean $0.5$.

\paragraph{Pretraining on ImageNet.}
We now show that it is possible to drastically improve the results by using more powerful embedding functions. Particularly, we follow the same procedure defined above, but we use different models pre-trained on ImageNet, instead of the respective training sets. The considered models are a ResNet50x1, ResNet50x2, and ResNet50x4 \cite{simclr}, all downloaded from the official repository.\footnote{https://github.com/google-research/simclr} In Table~\ref{tab:imagenet}, we report the results using the cosine similarity for all models. The results confirm the current trend in machine learning: the larger the model, the better the performance. Particularly, ResNet50x4 obtains the best results that we have achieved in this work with cosine similarity, with a huge improvement with respect to smaller models presented in Tab.~\ref{tab:results_stl}. This shows that the proposed method is general, and strongly benefits from large pre-trained models made available for transfer learning.

\begin{table*}[t]
    \caption{Percentage of wrongly retrieved memories on STL using pre-trained models on ImageNet.}
    \centering{\vspace*{-0.5ex}  
    	\resizebox{\textwidth}{!}{\begin{tabular}{@{}lcccccccc@{}}
    		\toprule
    		& Crop  & Mask & Color & Rotation & S\&P & Gauss $\mu =0.3$ & Gauss $\mu = 0.5$    \\
    		\toprule
    		 ResNet50x1 (STL10)  & $26.99\%$ & $64.12$\% & $0.01$\% & $0.01\%$ & $43.88$\% & $14.58$\% & $29.51\%$ \\ 
    		 
    		 ResNet50x2 (STL10)  & $19.95\%$  & $32.42\%$ & $0.0\%$ & $0.01\%$ & $23.40\%$ & $9.20\%$ &$16.2\%$ \\
    		 
    		 ResNet50x4 (STL10)  & $13.92\%$  & $11.74\%$ & $0.0\%$ & $0.0\%$  & $15.12\%$  & $4.22\%$ & $7.94\%$ \\
    		 
    		 
    		\midrule

    		 ResNet50x1 (ImageNet64)  & $24.82\%$ & $57.09$\% & $0.01$\% & $0.01\%$ & $36.42$\% & $12.55$\% & $27.74\%$ \\ 
    		 
    		 ResNet50x2 (ImageNet64)  & $17.64\%$  & $28.11\%$ & $0.0\%$ & $0.01\%$ & $21.71\%$ & $8.96\%$ &$14.99\%$ \\
    		 
    		 ResNet50x4 (ImageNet64)  & $13.01\%$  & $10.07\%$ & $0.0\%$ & $0.0\%$  & $13.63\%$  & $4.02\%$ & $7.18\%$ \\

            \bottomrule
    	\end{tabular}}}
	\label{tab:imagenet}
\end{table*}

\section{Fully-semantic Memory Model}

\begin{figure*}[t]
\includegraphics[width=1.0\textwidth]{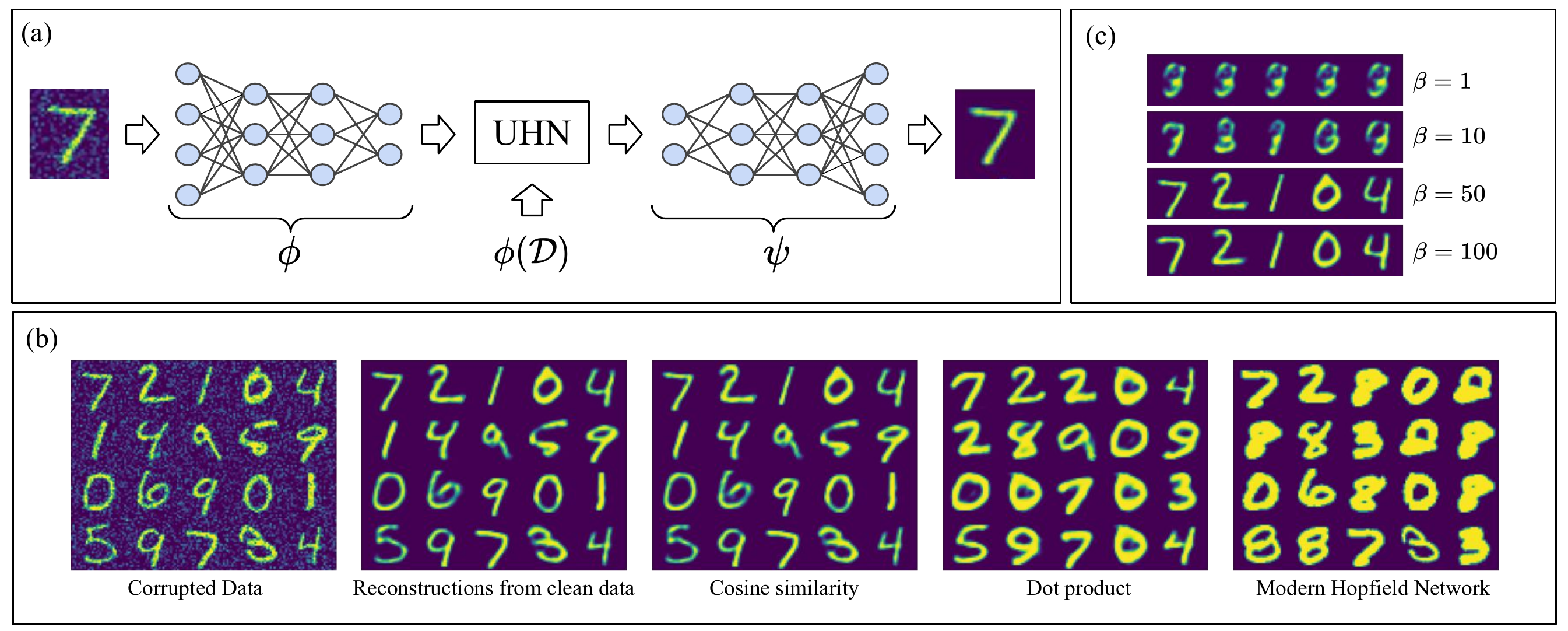}\vspace*{-0.5ex}\caption[short]{(a): Example of a fully-semantic memory model, where  $\phi$ and $\psi$ are the encoder and decoder parts of a trained autoencoder, and the goal is to retrieve an MNIST image given a corrupted version. (b)  Retrieved images when provided with a corrupted version of the first $20$ images of the MNIST test set with Gaussian noise of mean $0$ and variance $0.2$ (left). The best result is obtained with the cosine similarity, identical to the original retrievals of the autoencoder when provided with clean data. (c)~Examples of retrievals with the cosine similarity when varying the temperature constant $\beta$. }
\label{fig:shn}\vspace*{-0.5ex}\end{figure*}

From the biological perspective, the family of memory models introduced in the previous section is implausible, as it stores exact copies of the dataset in memory instead of low-dimensional representations. In fact, our brain poorly performs when it comes to exact retrievals, but it is excellent in recalling conceptual memories \cite{pineda2021entropic,rugg2013brain,wagner2005parietal}. Here, we provide a memory model that, on the one hand, is coherent with the biological constraints, and on the other hand, is more memory-efficient. The main drawback, however, is the inability of not retrieving memories exactly, often useful in practical tasks. As both scoring and retrievals are computed in a low-dimensional embedding space, we call this family of models \emph{fully-semantic memory model}. 

Note that both the score and the projection function defined in the previous section require access to a dataset $\mathcal D$. To overcome this, we need two functions $\phi$ and $\psi$, where $\phi$ is conceptually similar to the ones used for the semantic memory model, as it again maps data points to a low-dimensional embedding space $\mathbb{R}^e$, and $\psi$ is a generative function that follows the inverse path of mapping from the embedding space back to images. A  formal definition is as follows.
\begin{definition}
Given a dataset $\mathcal{D} = \{\bar x_i\}_{i \leq N}$, a feature map $\phi: \mathbb{R}^d \longrightarrow \mathbb{R}^e$, and a generative map $\psi: \mathbb{R}^e \longrightarrow \mathbb{R}^d$, a \emph{fully-semantic memory model} is a function ${\mu_{\phi(\mathcal{D})}: \mathbb{R}^d \longrightarrow \mathbb{R}^d}$ such that:
\begin{enumerate}
    \item $\mu_{\phi(\mathcal{D})}$ admits the decomposition $\mu_{\phi(\mathcal{D})} = \psi \circ \pi_{\phi(\mathcal{D})} \circ \alpha \circ \kappa_{\phi(\mathcal{D})} \circ \phi$,

    \item the map $\pi_{\phi(\mathcal{D})} \circ \alpha \circ \kappa_{\phi(\mathcal{D})} $ is a universal Hopfield network on the embedded dataset $\phi(\mathcal{D})$.
\end{enumerate} 
\end{definition}
Note that the dataset is not stored, but only its embeddings are. If the dimensionality of the embedding space is significantly smaller than the dimensionality of the data, then this results in significant memory savings. However, also the parametric functions $\phi$ and $\psi$ have to be stored, and hence the effective advantage in terms of memory is a tradeoff between these two quantities.
%


\paragraph{Learning $\phi$ and $\psi$.} To make the retrieval of the fully-semantic model effective, we need the functions $\phi$ and $\psi$ to be meaningful. This means that they  again have to be pre-trained on a dataset that has similar features to the ones that we want to store. We will now show an example on a small autoencoder, i.e., a multi-layer perceptron trained to generate the same data point used as an input. The distinguishing characteristic of an autoencoder is the presence of a \emph{bottleneck} layer,  much smaller than the input layer, which is required to prevent the network from simply learning the identity mapping. The sequence of layers that maps the input to the bottleneck layer is called \emph{encoder}; the remaining part, which maps the output of the encoder back to the input space is called \emph{decoder}. We consider the functions $\phi$ and $\psi$ to be the trained encoder and  decoder, respectively. A sketch of this network is shown in Fig.~\ref{fig:shn}(a). 

\paragraph{Set-Up.} The task that we tackle now is a standard one in the associative memory domain: we present the model with a corrupted version of an image that it has stored in memory as a key, and check whether the model is able to retrieve an image that is semantically equivalent to the original one. As a consequence, the results that we present in this section are purely qualitative, as it does not make sense to score images based on how similar they are to the original with respect to a distance on the pixel space. 
To learn the functions $\phi$ and $\psi$, we  trained an autoencoder to generate images of the training dataset, composed of $60000$ images. Then, we perform associative memory tasks on the test set, composed of $10000$ images. To do that, we  first saved the embedding of the test set (every embedding has the dimension $12$), and then corrupted every image with Gaussian noise. As similarity functions, we  tested the dot product and the cosine similarity, and as separation functions, we used the softmax with different inverse temperatures $\beta$. For completeness, we have also reported the reconstructions of MHNs, by using the dot product as a similarity function. In both cases, we have not performed any normalization before scoring the similarities.

\paragraph{Implementation Details.} The autoencoder has $8$ layers of dimension $784,64,32,16,12,16,32,64,784$, and was trained with a ReLU activation, learning rate of $0.001$, and a batch size of $250$ for $300$ epochs. The corruption used is simple Gaussian noise with mean $0$ and variance $0.2$. Training the functions $\phi$ and $\psi$ (hence, the autoencoder) takes approximately $5$ minutes on an RTX Titan.
Note that the experiments proposed for fully-semantic models are not to be considered for practical applications, as we have used a simple and deterministic generative model. 

\paragraph{Results.} Representations of corrupted keys, as well as the retrieved constructive memories, are given in Fig.~\ref{fig:shn}(b). Particularly, the reconstructions show that the model is able to correctly retrieve memories in the embedding space, even when the cardinality of the dataset is large. However, the retrieval is not perfect, and sporadic errors may occur. These results can be improved, and scaled up to more complex datasets, by using more complex encoders and decoders. In terms of functions used, the cosine similarity outperforms the dot product, and the softmax with large inverse temperatures $(\beta \leq 50)$ is needed for one-shot retrievals, as shown in Fig.~\ref{fig:shn}(c). In fact, a softmax separation function with a small temperature is not enough to discriminate between different stored data points when performing one-shot retrievals.

\section{Related Work}

While using the same networks as in several computer vision tasks, the final goal of our work is to perform memory tasks, and is hence mostly related with the associative memory literature. The first model of this kind, called the \emph{learnmatrix} \cite{Steinbuch61}, dates back to 1961, and was built using the hardware properties of ferromagnetic circuits. The first two influential computational models, however, are the Hopfield network \cite{Hopfield82,Hopfield84} and sparse-distributed memory models \cite{kanerva1988sparse}. The first emulates the dynamics of a spin-glass system, and the second was born as a computational model of our long-term memory system. In recent years, associative memory models have re-gained popularity, as their literature is increasingly intersecting that of deep learning. A variation of Hopfield networks with polynomial capacity has been introduced to perform classification tasks \cite{krotov16}, and a sequential result showed that this capacity can be made exponential with a simple change of activation function \cite{Demircigil17}. However, these models were used to perform classification tasks also due to their limitation in dealing only with discrete vectors. The generalization to continuous valued vectors has been developed several years later \cite{ramsauer21}. There is also a line of research that uses associative memory models mixed with deep architectures, such as deep associative neural networks \cite{LIU19}, which augment the storage and retrieval mechanism of dense Hopfield networks using deep belief networks \cite{hinton2009deep}, and generative predictive coding networks \cite{salvatori2021associative, salvatori2023brain}, which rely on the theory of predictive coding to store and retrieve images. Recent lines of works have also focused in implementing \emph{forget} operations, to remove stored memories that are not needed anymore \cite{yoo2022bayespcn, ota2023learning}.

While many works primarily focus on retrieval tasks, recent ones have also used associative memory models to study and understand the popular transformer architecture \cite{Vaswani17}. It has in fact been shown that the attention mechanism is  a special formulation of modern continuous-state Hopfield networks \cite{ramsauer21}, and that their dynamics can also be approximated by a modern formulation of sparse-distributed memory models \cite{bricken2021attention}. A similar result has been proven for the fully MLP architecture \cite{mixer}, able to achieve excellent results in classification tasks despite only using fully connected layers. 

\section{Conclusion}

In this work, we have addressed the problem of storing and retrieving natural data  such as colored images in associative memory models. First, we have discussed the problem of computing similarities on the pixel space, which  creates a mismatch between human and machine performance when it comes to associate similar stored data points. Due to the fact that modern associative memory models compute simple similarity scores on raw pixels, it is in fact possible to simply rotate or translate an image to trick modern memory models. The same transformations, however, would not be able to trick a human judge. To address this mismatch, we have defined two associative memory models that compute similarity scores in an embedding space, allowing to perform associative  memory tasks in scenarios where corruptions do not alter the conceptual content of the stored data points.

In terms of generality of the considered benchmarks, we have tested against an associative memory model that is a generalization of most of the models present in the literature, the universal Hopfield network. In detail, it
is a generalization of modern Hopfield networks, continuous state Hopfield networks, as well as Kanerva associative memories. Hence, we believe that our analysis is rich enough, as
it shows how the performance is sometimes orders of magnitude better. In terms of architecture considered, we have used ResNets, as they are both the most powerful pre-trained models available with contrastive loss, as well as the ones expected to achieve a better performance. Hence, we expect the results of almost any other class of models to be worse than the ones obtained in this work. However, our method is highly generalizable: given any
state-of-the-art (SOTA) memory model X, we can apply our embedding function to enhance X’s retrieval performance for natural images while significantly increasing capacity. This
generalizability eliminates the need to test against every individual model, as our method naturally improves performance
by leveraging the quality of the embedding from a large pretrained ResNet.

As embedding models, we have used neural networks trained with a contrastive loss. As this is a popular method in the modern literature, it is easy to find pre-trained models suitable for a given task, freeing the user from the burden of training one from scratch. Training your own contrastive model, however, has an interesting advantage for some practical applications, where original data points are often faced with the same kind of corruptions. One example is that of adversarial attacks: let us assume our memory model gets always tricked by one kind of corruption, it is now possible to collect multiple examples of this corruption, and feed it in the contrastive loss using them as data augmentation. This would enforce the model to group together corrupted versions of the same data point, where the corruption is the same one that will be faced by the dataset. The second model that we propose has the goal of making the model lighter and more plausible, as well as generating images similar, but not identical, to the stored ones. It is a fully semantic model, which performs both similarities and reconstructions in the embedding space. We have proposed simple experiments on an autoencoder trained on  MNIST. Applications in practice would need more powerful generative models, picked according to the needed task and data. 

\section*{Acknowledgments}
Thomas Lukasiewicz
was supported by the Alan Turing Institute under
the UK EPSRC grant EP/N510129/1, the AXA Research Fund, and the EU TAILOR grant 952215.
Rafal Bogacz was supported by the UK BBSRC grant BB/S006338/1, and the UK MRC grant MC\textunderscore UU\textunderscore 00003/1.

\bibliography{references}

\begin{thebibliography}{10}

\bibitem{Barron20}
Helen~C. Barron, Ryszard Auksztulewicz, and Karl Friston, `Prediction and
  memory: A predictive coding account', {\em Progress in Neurobiology}, {\bf
  192},  101821, (2020).

\bibitem{bricken2021attention}
Trenton Bricken and Cengiz Pehlevan, `Attention approximates sparse distributed
  memory', {\em Advances in Neural Information Processing Systems}, {\bf 34},
  (2021).

\bibitem{chen2020simple}
Ting Chen, Simon Kornblith, Mohammad Norouzi, and Geoffrey Hinton, `A simple
  framework for contrastive learning of visual representations', in {\em
  International Conference on Machine Learning}. PMLR, (2020).

\bibitem{simclr}
Ting Chen, Simon Kornblith, Mohammad Norouzi, and Geoffrey Hinton, `A simple
  framework for contrastive learning of visual representations', {\em
  Proceedings of the 37th International Conference on Machine Learning},
  (2020).

\bibitem{chen2020big}
Ting Chen, Simon Kornblith, Kevin Swersky, Mohammad Norouzi, and Geoffrey
  Hinton, `Big self-supervised models are strong semi-supervised learners',
  {\em 34th Conference on Neural Information Processing Systems, NeurIPS},
  (2020).

\bibitem{stl10}
Adam Coates, Andrew Ng, and Honglak Lee, `An analysis of single-layer networks
  in unsupervised feature learning', in {\em Proceedings of the 14th
  International Conference on Artificial Intelligence and Statistics}, (2015).

\bibitem{Demircigil17}
Mete Demircigil, Judith Heusel, Matthias Löwe, Sven Upgang, and Franck Vermet,
  `On a model of associative memory with huge storage capacity', {\em Journal
  of Statistical Physics}, {\bf 168}, (2017).

\bibitem{he2016deep}
Kaiming He, Xiangyu Zhang, Shaoqing Ren, and Jian Sun, `Deep residual learning
  for image recognition', in {\em Proceedings of the IEEE Conference on
  Computer Vision and Pattern Recognition}, (2016).

\bibitem{hinton2009deep}
Geoffrey~E. Hinton, `Deep belief networks', {\em Scholarpedia}, {\bf 4}(5),
  5947, (2009).

\bibitem{Hopfield82}
John~J. Hopfield, `Neural networks and physical systems with emergent
  collective computational abilities', {\em Proceedings of the National Academy
  of Sciences}, {\bf 79}, (1982).

\bibitem{Hopfield84}
John~J. Hopfield, `Neurons with graded response have collective computational
  properties like those of two-state neurons', {\em Proceedings of the National
  Academy of Sciences}, {\bf 81}, (1984).

\bibitem{kanerva1988sparse}
Pentti Kanerva, {\em Sparse Distributed Memory}, MIT Press, 1988.

\bibitem{Krizhevsky2012}
Alex Krizhevsky, Ilya Sutskever, and Geoffrey~E. Hinton, `{ImageNet}
  classification with deep convolutional neural networks', in {\em 26th Annual
  Conference on Neural Information Processing Systems (NIPS) 2012}, (2012).

\bibitem{krotov16}
Dmitry Krotov and John~J. Hopfield, `Dense associative memory for pattern
  recognition', in {\em Advances in Neural Information Processing Systems},
  (2016).

\bibitem{Krotov21}
Dmitry Krotov and John~J. Hopfield, `Large associative memory problem in
  neurobiology and machine learning', in {\em International Conference on
  Learning Representations}, (2021).

\bibitem{LIU19}
Jia Liu, Maoguo Gong, and Haibo He, `Deep associative neural network for
  associative memory based on unsupervised representation learning', {\em
  Neural Networks}, {\bf 113},  41--53, (2019).

\bibitem{millidge2022universal}
Beren Millidge, Tommaso Salvatori, Yuhang Song, Thomas Lukasiewicz, and Rafal
  Bogacz, `Universal {H}opfield networks: A general framework for single-shot
  associative memory models', {\em arXiv:2202.04557}, (2022).

\bibitem{ota2023learning}
Toshihiro Ota, Ikuro Sato, Rei Kawakami, Masayuki Tanaka, and Nakamasa Inoue,
  `Learning with partial forgetting in modern {H}opfield networks', in {\em
  International Conference on Artificial Intelligence and Statistics}, pp.
  6661--6673. PMLR, (2023).

\bibitem{pineda2021entropic}
Luis~A. Pineda, Gibr{\'a}n Fuentes, and Rafael Morales, `An entropic
  associative memory', {\em Scientific Reports}, {\bf 11}(1),  1--15, (2021).

\bibitem{ramsauer21}
Hubert Ramsauer, Bernhard Sch{\"a}fl, Johannes Lehner, Philipp Seidl, Michael
  Widrich, Lukas Gruber, Markus Holzleitner, Thomas Adler, David Kreil,
  Michael~K. Kopp, G{\"u}nter Klambauer, Johannes Brandstetter, and Sepp
  Hochreiter, `Hopfield networks is all you need', in {\em International
  Conference on Learning Representations}, (2021).

\bibitem{rugg2013brain}
Michael~D. Rugg and Kaia~L. Vilberg, `Brain networks underlying episodic memory
  retrieval', {\em Current Opinion in Neurobiology}, {\bf 23}(2),  255--260,
  (2013).

\bibitem{salvatori2021associative}
Tommaso Salvatori, Yuhang Song, Yujian Hong, Lei Sha, Simon Frieder, Zhenghua
  Xu, Rafal Bogacz, and Thomas Lukasiewicz, `Associative memories via
  predictive coding', {\em Advances in Neural Information Processing Systems},
  {\bf 34}, (2021).

\bibitem{Steinbuch61}
Karl Steinbuch, `Die {L}ernmatrix', {\em Kybern.}, {\bf 1}(1),  36--45, (1961).

\bibitem{mixer}
Ilya~O. Tolstikhin, Neil Houlsby, Alexander Kolesnikov, Lucas Beyer, Xiaohua
  Zhai, Thomas Unterthiner, Jessica Yung, Andreas Steiner, Daniel Keysers,
  Jakob Uszkoreit, Mario Lucic, and Alexey Dosovitskiy, `{MLP-Mixer: An
  all-MLP} architecture for vision', {\em CoRR}, {\bf abs/2105.01601}, (2021).

\bibitem{Vaswani17}
Ashish Vaswani, Noam Shazeer, Niki Parmar, Jakob Uszkoreit, Llion Jones,
  Aidan~N. Gomez, Lukasz Kaiser, and Illia Polosukhin, `Attention is all you
  need', in {\em Advances in Neural Information Processing Systems 30}, (2017).

\bibitem{wagner2005parietal}
Anthony~D. Wagner, Benjamin~J. Shannon, Itamar Kahn, and Randy~L. Buckner,
  `Parietal lobe contributions to episodic memory retrieval', {\em Trends in
  Cognitive Sciences}, {\bf 9}(9),  445--453, (2005).

\bibitem{yoo2022bayespcn}
Jinsoo Yoo and Frank Wood, `{BayesPCN: A} continually learnable predictive
  coding associative memory', {\em Advances in Neural Information Processing
  Systems}, {\bf 35},  29903--29914, (2022).

\end{thebibliography}

\end{document}